\documentclass[10pt,twocolumn,letterpaper]{article}

\usepackage{iccv}              % To produce the CAMERA-READY version
\usepackage{graphicx}
\usepackage{colortbl} % 需要 colortbl 宏包支持
\usepackage{xcolor}   % 额外的颜色支持
\usepackage{enumitem}
\usepackage{float}
\usepackage{tikz}
\usepackage{booktabs}
\usepackage{multirow}
\usepackage{marvosym} % 引入 wasysym 宏包

\newcommand{\name}{MaTVLM}

\definecolor{iccvblue}{rgb}{0.21,0.49,0.74}
\usepackage[pagebackref,breaklinks,colorlinks,allcolors=iccvblue]{hyperref}

\title{MaTVLM: Hybrid Mamba-Transformer for Efficient Vision-Language Modeling}

\author{Yingyue Li$^{1}$ 
  \ \ \ \  Bencheng Liao$^{2,1}$ 
  \ \ \ \  Wenyu Liu$^{1}$ 
  \ \ \ \  Xinggang Wang$^{1, \textrm{\Letter}}$\\
$^{1}$ School of EIC, Huazhong University of Science \& Technology\\
$^{2}$ Institute of Artificial Intelligence, Huazhong University of Science \& Technology
}

\begin{document}
% \maketitle
\twocolumn[
\maketitle
\vskip -0.3in
{
\vspace{-0.5cm}
\begin{center}
    \captionsetup{type=figure}
    \includegraphics[width=1.0\textwidth]{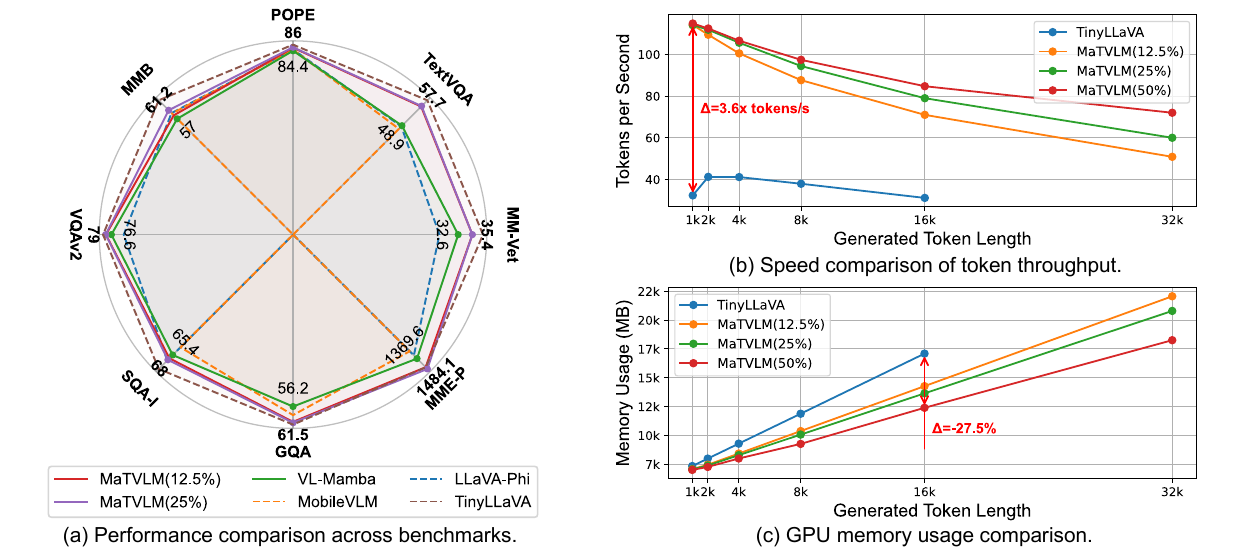}
    % \vspace{-.1 in}
    \captionof{figure}{
        \textbf{Comprehensive comparison of our \name.} 
        \textbf{(a)} Performance comparison across multiple benchmarks. Our \name~achieves competitive results with the teacher model TinyLLaVA, surpassing existing VLMs with similar parameter scales, as well as Mamba-based VLMs.
        \textbf{(b)} Speed Comparison of Token Throughput. Tokens generated per second for different token lengths. Our \name~achieves a $3.6\times$ speedup compared to the teacher model TinyLLaVA.
        \textbf{(c)} GPU Memory Usage Comparison. A detailed comparison of memory usage during inference for different token lengths, highlighting the optimization advantages with a 27.5\% reduction in usage for our \name~over TinyLLaVA.
    }
    \label{fig:teaser}
\end{center} %
}]
\let\thefootnote\relax\footnotetext{$^\boxtimes$ Corresponding author: \texttt{xgwang@hust.edu.cn}.}

\begin{abstract}

With the advancement of RNN models with linear complexity, the quadratic complexity challenge of transformers has the potential to be overcome. Notably, the emerging Mamba-2 has demonstrated competitive performance, bridging the gap between RNN models and transformers. 
However, due to sequential processing and vanishing gradients, RNN models struggle to capture long-range dependencies, limiting contextual understanding. This results in slow convergence, high resource demands, and poor performance on downstream understanding and complex reasoning tasks. 
In this work, we present a hybrid model \name~
by substituting a portion of the transformer decoder layers in a pre-trained VLM with Mamba-2 layers. Leveraging the inherent relationship between attention and Mamba-2, we initialize Mamba-2 with corresponding attention weights to accelerate convergence. Subsequently, we employ a single-stage distillation process, using the pre-trained VLM as the teacher model to transfer knowledge to the \name, further enhancing convergence speed and performance. Furthermore, we investigate the impact of differential distillation loss within our training framework.
We evaluate the \name~on multiple benchmarks, demonstrating competitive performance against the teacher model and existing VLMs while surpassing both Mamba-based VLMs and models of comparable parameter scales. Remarkably, the \name~achieves up to $3.6\times$ faster inference than the teacher model while reducing GPU memory consumption by 27.5\%, all without compromising performance. Code and models are released at \url{https://github.com/hustvl/MaTVLM}.

\end{abstract}    
\section{Introduction}
   
Large vision-language models (VLMs) have rapidly advanced in recent years~\cite{bai2023qwenvl, wang2024qwen2vl, chen2024internvl, zhou2024tinyllava, liu2024llava, liu2024improvedllava, chen2024internvl2, huang2024mini}. VLMs are predominantly built on transformer architecture. However, due to the quadratic complexity of transformer with respect to sequence length, VLMs are computationally intensive for both training and inference. 
Recently, several RNN models~\cite{yang2023gated, gu2023mamba, sun2023retentive, de2024griffin, dao2024mamba2} have emerged as potential alternatives to transformer, offering linear scaling with respect to sequence length. Notably, Mamba~\cite{gu2023mamba, dao2024mamba2} has shown exceptional performance in long-range sequence tasks, surpassing transformer in computational efficiency. 

Several studies~\cite{qiao2024vlmamba, zhao2024cobra, huang2024mlmamba, vim, vig, dig} have explored integrating Mamba architecture into VLMs by replacing transformer-based large language models (LLMs) with Mamba-based LLMs. These works have demonstrated competitive performance while achieving significant gains in inference speed. However, several limitations are associated with these approaches:
\textbf{(1)} Mamba employs sequential processing, which limits its ability to capture global context compared to transformer, thereby restricting these VLMs' performance in complex reasoning and problem-solving tasks~\cite{wen2024rnns,yang2024efficient};
\textbf{(2)} The sequential nature of Mamba results in inefficient gradient propagation during long-sequence training, leading to slow convergence when training VLMs from scratch. As a result, the high computational cost and the large amount of training data required become significant bottlenecks for these VLMs;
\textbf{(3)} The current training scheme for these VLMs is complex, requiring multi-stage training to achieve optimal performance. This process is both time-consuming and computationally expensive, making it difficult to scale Mamba-based VLMs for broader applications.

To address the aforementioned issues, we propose a novel \textbf{Ma}mba-\textbf{T}ransformer \textbf{V}ision-\textbf{L}anguage \textbf{M}odel (\name) that integrates Mamba-2 and transformer components, striking a balance between computational efficiency and overall performance. \textbf{Firstly}, attention and Mamba are inherently connected, removing the softmax from attention transforms it into a linear RNN, revealing its structural similarity to Mamba. We will analyze this relationship in detail in Sec.~\ref{sec:hybrid}. \textbf{Furthermore}, studies applying Mamba to large language models (LLMs)~\cite{waleffe2024empirical, wang2024mambainllama} have demonstrated that models hybridizing Mamba outperform both pure Mamba-based and transformer-based models on certain tasks. Motivated by this connection and empirical findings, combining Mamba with transformer components presents a promising direction, offering a trade-off between improved reasoning capabilities and computational efficiency. Specifically, we adopt the TinyLLaVA~\cite{zhou2024tinyllava} as the base VLM and replace a portion of its transformer decoder layers with Mamba decoder layers while keeping the rest of the model unchanged.

To minimize the training cost of the \name~while maximizing its performance, we propose to distill knowledge from the pre-trained base VLM. \textbf{Firstly}, we initialize Mamba-2 with the corresponding attention's weights as mentioned in Sec.~\ref{sec:hybrid}, which is important to accelerate the convergence of Mamba-2 layers. \textbf{Moreover}, during distillation training, we employ both probability distribution and layer-wise distillation losses to guide the learning process, making only Mamba-2 layers trainable while keeping transformer layers fixed. \textbf{Notably}, unlike most VLMs that require complex multi-stage training, our approach involves a single-stage distillation process. 

Despite the simplified training approach, our model demonstrates comprehensive performance across multiple benchmarks, as illustrated in Fig.~\ref{fig:teaser}. It exhibits competitive results when compared to the teacher model, TinyLLaVA, and outperforms Mamba-based VLMs as well as other transformer-based VLMs with similar parameter scales. The efficiency of our model is further emphasized by a $3.6\times$ speedup and a 27.5\% reduction in memory usage, thereby confirming its practical advantages in real-world applications. These results underscore the effectiveness of our approach, providing a promising avenue for future advancements in model development and optimization.

In summary, this paper makes three significant contributions:
\begin{itemize}[leftmargin=0.5cm] 
   \item We propose a new hybrid VLM architecture \name~that effectively integrates Mamba-2 and transformer components, balancing the computational efficiency with high-performance capabilities.
   \item We propose a novel single-stage knowledge distillation approach for the Mamba-Transformer hybrid VLMs. By leveraging pre-trained knowledge, our method accelerates convergence, enhances model performance, and strengthens visual-linguistic understanding.
    \item We demonstrate that our approach significantly achieves a $3.6\times$ faster inference speed and a 27.5\% reduction in memory usage while maintaining the competitive performance of the base VLM. Moreover, it outperforms Mamba-based VLMs and existing VLMs with similar parameter scales across multiple benchmarks.
\end{itemize}

\section{Related Work}

\subsection{Efficient VLMs}

In recent years, efficient and lightweight VLMs have advanced significantly.
Several academic-oriented VLMs, such as TinyLLaVA-3.1B~\cite{zhou2024tinyllava}, MobileVLM-3B~\cite{chu2023mobilevlm}, and LLaVA-Phi~\cite{zhu2024llavaphi}  have been developed to improve efficiency. Meanwhile, commercially oriented models like Qwen2.5-VL-3B~\cite{bai2025qwen2.5}, InternVL2.5-2B~\cite{chen2024internvl2.5}, and others achieve remarkable performance by leveraging large-scale datasets with high-resolution images and long-context text.

Our work prioritizes efficiency and resource constraints over large-scale, commercially oriented training. Unlike previous approaches, by integrating Mamba-2~\cite{dao2024mamba2}, our method achieves competitive performance while significantly reducing computational demands, making it well-suited for deployment in resource-limited environments.

\subsection{Structured State Space Models}
Structured state space models (S4)~\cite{gu2021s4, gupta2022DSS, smith2022S5, ma2022mega, gu2023mamba, dao2024mamba2} scale efficiently in a linear manner with sequence length.
 Mamba\cite{gu2023mamba} introduces selective SSMs, while Mamba-2~\cite{dao2024mamba2} refines this by linking SSMs to attention variants, achieving 2–8× speedup and performance comparable to transformers.
Mamba-based VLMs~\cite{huang2024mlmamba, qiao2024vlmamba, zhao2024cobra, zou2025omnimambaefficientunifiedmultimodal} primarily replace the transformer-based large language models (LLMs) entirely with the pre-trained Mamba-2 language model, achieving both competitive performance and enhanced computational efficiency.

Our work innovatively integrates Mamba and transformer within VLMs, combining their strengths rather than entirely replacing transformers with Mamba-2. By adopting a hybrid approach and introducing a single-stage distillation strategy, we enhance model expressiveness, improve efficiency, and achieve superior performance over previous Mamba-based VLMs while maintaining computational efficiency for practical deployment.

\subsection{Hybrid Mamba and Transformer}
Recent works, such as MambaInLlama~\cite{wang2024mambainllama} and MOHAWK~\cite{bick2025transformerstossm}, demonstrate the effectiveness of hybrid Mamba-Transformer architectures in LLMs, achieving notable improvements in efficiency and performance. Additionally, MambaVision~\cite{hatamizadeh2024mambavision} extends this hybrid approach to vision models, introducing a Mamba-Transformer-based backbone that excels in image classification and other vision-related tasks, showcasing the potential of integrating SSMs with transformers.

Unlike previous studies on LLMs or vision backbones, our work extends the hybrid Mamba-Transformer to VLMs and design a concise architecture with an efficient single-stage distillation strategy, enhancing convergence, reducing inference time, and lowering memory consumption for practical deployment.

\subsection{Knowledge Distillation}
More recently, knowledge distillation for LLMs has gained attention~\cite{Gu_Dong_Wei_Huang_2023_KDLLM,Wen_Li_Du_Mou_sequenceKD,Agarwal_2024_Generated,wang2020minilm,jha2023largeteacherless}, while studies on VLM distillation remain limited~\cite{Fang2021CompressVL,Wang2022encoderVL,xu2024llavadi}. DistillVLM~\cite{Fang2021CompressVL} uses MSE loss to align attention and feature maps, MAD~\cite{Wang2022encoderVL} aligns visual and text tokens, and LLAVADI~\cite{xu2024llavadi} highlights the importance of joint token and logit alignment.

Building on these advancements, we integrate knowledge distillation into a hybrid Mamba-Transformer framework with a single-stage distillation strategy to transfer knowledge from a transformer-based teacher model. This improves convergence, enhances performance, and reduces computational costs for efficient VLM deployment.

 \section{Method}
 Large vision-language models (VLMs) process longer sequences than LLMs, resulting in slower training and inference. As previously mentioned, Mamba-2 architecture exhibits linear scaling and offers significantly higher efficiency compared to transformer. To leverage these advantages, we propose a hybrid VLM architecture \name~that integrates Mamba-2 and transformer components, aiming to balance computational efficiency with optimal performance.

 \subsection{Mamba Preliminaries}
 Mamba~\cite{gu2023mamba} is mainly built upon the structured state-space sequence models (S4) as in Eq.~\ref{eq: s4}, which are a recent development in sequence modeling for deep learning, with strong connections to RNNs, CNNs, and classical state space models.
 \begin{align}
    \label{eq: s4}
    h_t &= \overline{\mathbf{A}} h_{t-1} + \overline{\mathbf{B}} x_t, \notag \\
    y_t &= \mathbf{C}^\top h_t.
 \end{align}
 
Mamba has introduced the selective state space models (Selective SSMs), as shown in Eq.~\ref{eq: mamba2}, unlike the standard linear time-invariant (LTI) formulation~\ref{eq: s4}, enables the ability to selectively focus on or ignore inputs at each timestep. Its performance has been shown to surpass LTI SSMs on information-rich tasks such as language processing, especially when the state size \( N \) grows, allowing it to handle a larger capacity of information.
\begin{align}
\label{eq: mamba2}
h_t &= \mathbf{A}_t h_{t-1} + \mathbf{B}_t x_t, \notag \\
y_t &= \mathbf{C}_t^\top h_t.
\end{align}

Mamba-2~\cite{dao2024mamba2} advances Mamba’s selective SSMs by introducing the state-space duality (SSD) framework, which establishes a theoretical link between SSMs and various attention mechanisms through different decompositions of structured semi-separable matrices. Leveraging this framework, Mamba-2 achieves $2–8\times$ faster computation while maintaining competitive performance with transformers.

\begin{figure*}[htbp]
    \begin{center}
       \includegraphics[width=1.0\linewidth]{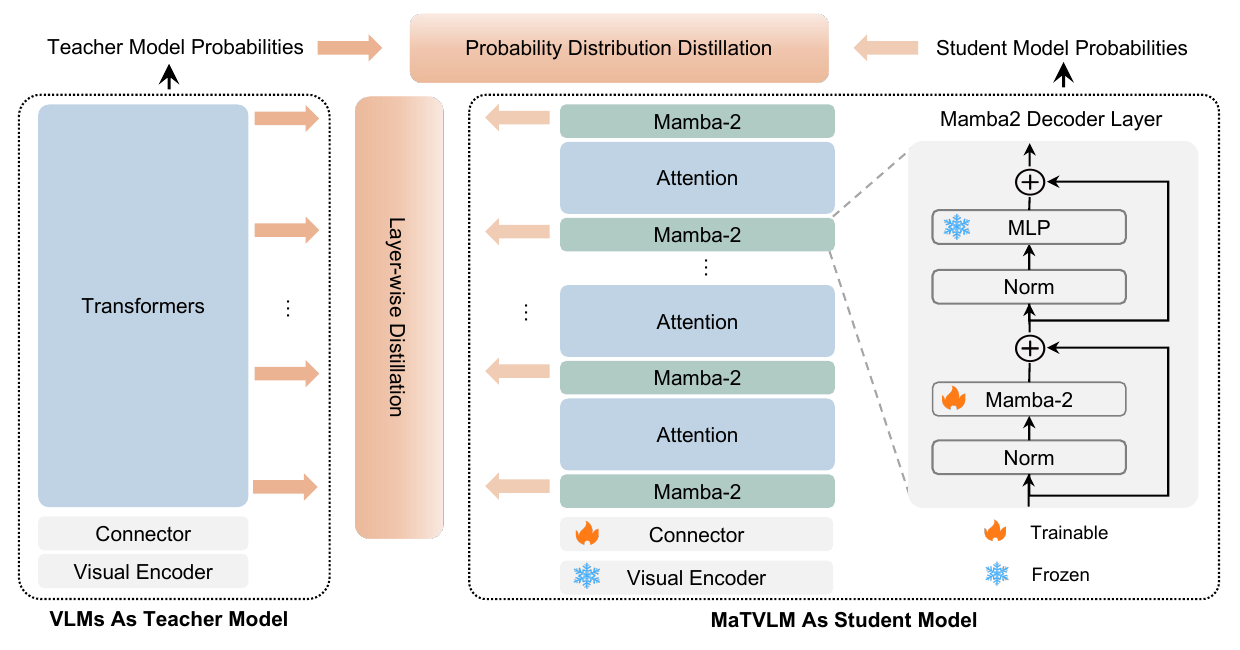}
    \end{center}
    \vspace{-0.7cm}
       \caption{The proposed \name~integrates both Mamba-2 and transformer components. The model consists of a vision encoder, a connector, and a language model same as the base VLM. The language model is composed of both transformer decoder layers and Mamba-2 decoder layers, where Mamba-2 layers replace only attention in transformer layers, while the other components remain unchanged. The model is trained using a knowledge distillation approach, incorporating probability distribution and layer-wise distillation loss. During the distillation training, only Mamba-2 layers and the connector are trainable, while transformer layers remain fixed. 
       }
    % \vspace{-0.3cm}
    \label{fig:model}
    \end{figure*}

 \subsection{Hybrid Attention with Mamba for VLMs}
 \label{sec:hybrid}
 
 As shown in Fig.~\ref{fig:model}, the \name~is built upon the pre-trained VLMs, comprising a vision encoder, a connector, and a language model. The language model originally consists of transformer decoder layers, some of which are replaced with Mamba-2 decoder layers in our model. This replacement modifies only attention to Mamba-2 while leaving other components unchanged. 
 Based on the configured proportions (e.g., 12.5\%, 25\%) of Mamba-2 decoder layers, we distribute them at equal intervals. Given that Mamba-2 shares certain connections with attention, some weights can be partially initialized from the original transformer layers, as detailed below.
 
 Formally, for the $x_t$ in then  input sequence \( x = [x_1, x_2, \dots, x_n] \), attention in a transformer decoder layer is defined as:
 \begin{align}
    \mathbf{Q}_t &= x_t W_Q,\  
    \mathbf{K}_t = x_t W_K,\ 
    \mathbf{V}_t = x_t W_V, \notag\\
    y_n &= \sum_{t=1}^{n} \text{softmax}\left(\frac{\mathbf{Q}_n \mathbf{K}_t^\top }{\sqrt{d}}\right) \mathbf{V}_t, 
    \label{eq:attention}
 \end{align}
 where $d$ is the dimension of the input embedding, and $W_Q$, $W_K$, and $W_V$ are learnable weights. 
 
 When removing the softmax operation in Eq.~\ref{eq:attention}, the attention becomes:
 \begin{align}
    \label{eq:linear attention}
    y_n &= \sum_{t=1}^{n} \frac{\mathbf{Q}_n \mathbf{K}_t^\top }{\sqrt{d}} \mathbf{V}_t = \frac{\mathbf{Q}_n}{\sqrt{d}}\sum_{t=1}^{n} \mathbf{K}_t^\top  \mathbf{V}_t \notag \\
            &= \frac{\mathbf{Q}_n}{\sqrt{d}}\sum_{t=1}^{n} \mathbf{K}_t^\top  W_V x_t.
 \end{align}
 
 The above results can be reformulated in the form of a linear RNN as follows:
 \begin{align}
    \label{eq:linear rnn}
    h_n &= h_{n-1} + \mathbf{K}_n^\top  W_V x_n, \notag \\
    y_n &= \frac{\mathbf{Q}_n}{\sqrt{d}}h_n.
 \end{align}
 
 Comparing Eq.~\ref{eq:linear rnn} with the Eq.~\ref{eq: mamba2}, we can observe the following mapping relationships between them:
 \begin{align}
   \quad x_t = W_V x_t, \quad \mathbf{B}_t &= W_K x_t, \quad \mathbf{C}_t = W_Q x_t.
 \end{align}
 
\begin{figure}[htbp]
   \begin{center}
      \includegraphics[width=1.0\linewidth]{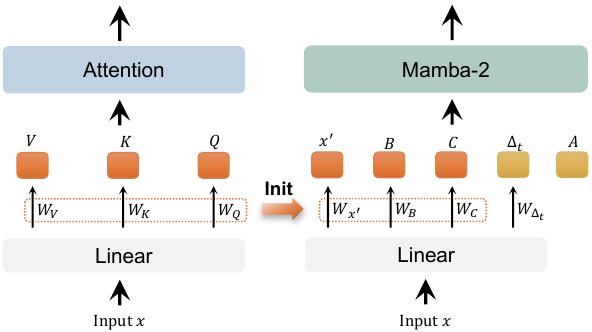}
   \end{center}
    \vspace{-0.5cm}
      \caption{ We initialize certain weights of Mamba-2 from attention based on their correspondence. Specifically, the linear weights of \( x, B, C \) in Mamba-2 are initialized from the linear weights of \( V, K, Q \) in the attention mechanism. The remaining parameters, including \( \Delta_t \) and \( A \), are initialized randomly.
      }
    % \vspace{-0.3cm}
   \label{fig:init}
   \end{figure}

 Consequently, we initialize the aforementioned weights of Mamba-2 layers with the corresponding weights from transformer layers as shown in Fig.~\ref{fig:init}, while the remaining weights are initialized randomly. Apart from Mamba-2 layers, all other weights remain identical to those of the original transformer.

 \label{sec:distillation}
 \subsection{Knowledge Distilling Transformers into Hybrid Models}
 To further enhance the performance of the \name, we propose a knowledge distillation method that transfers knowledge from transformer layers to Mamba-2 layers. We use a pre-trained VLM as the teacher model and our \name~as the student model. We will introduce the distillation strategies in the following.

    \vspace{-0.2cm}
 \paragraph{Probability Distribution Distillation}
 First, our goal is to minimize the distance between probability distributions of the models, just the logits output by the models before applying the softmax function. This approach is widely adopted in knowledge distillation, as aligning the output distributions of the models allows the student model to gain a more nuanced understanding from the teacher model's prediction. To achieve this, we use the kullback-leibler (KL) divergence with a temperature scaling factor as the loss function. The temperature factor adjusts the smoothness of the probability distributions, allowing the student model to capture finer details from the softened distribution of the teacher model. The loss function is defined as follows:
 \begin{align}
 L_{\text{prob}} &= T^2 \cdot \mathrm{KL}(\mathbf{P}_t \parallel \mathbf{P}_s) \notag \\
               &= T^2 \cdot \sum_{i} \mathbf{P}_{t}(i) \log\left(\frac{\mathbf{P}_{t}(i)}{\mathbf{P}_{s}(i)}\right),
 \end{align}
 The softened probabilities \( P_t(i) \) and \( P_s(i) \) are calculated by applying a temperature-scaled softmax function to the logits of the teacher and student models, respectively:
 \begin{align}
 \mathbf{P}_t &= \frac{\exp(z_t / T)}{\sum_j \exp(z_{t,j} / T)}, \notag \\
 \mathbf{P}_s &= \frac{\exp(z_s / T)}{\sum_j \exp(z_{s,j} / T)},
 \end{align}
 where \( T \) is the temperature scaling factor, a higher temperature produces softer distributions, \( z_t \) is the logit (pre-softmax output) from the teacher model, and \( \hat{z}_s \) is the corresponding logit from the student model. 
  
    \vspace{-0.2cm}

 \paragraph{Layer-wise Distillation}
 Moreover, to ensure that each Mamba layer in the student model aligns with its corresponding layer in the teacher model, we adopt a layer-wise distillation strategy. Specifically, this approach minimizes the L2 norm between the outputs of Mamba layers in the student model and the corresponding transformer layers in the teacher model when provided with the same input. These inputs are generated from the previous layer of the teacher model, ensuring consistency and continuity of context. By aligning intermediate feature representations, the student model can more effectively replicate the hierarchical feature extraction process of the teacher model, thereby enhancing its overall performance.
 Assume the Mamba layers' position in the student model is  \( l = [l_1, l_2, \dots, l_m] \).
 The corresponding loss function for this alignment is defined as:
 \begin{align}
    \label{eq:layer distillation}
    &\mathcal{L}_{\text{layer}} = \sum_{i=1}^{m} \left\| \mathbf{T}_{l_i}(x) - \mathbf{S}_{l_i}(x) \right\|_2, 
 \end{align}
where \( \mathbf{T}_{l_i}(x) \) and \( \mathbf{S}_{l_i}(x) \) represent the outputs of the teacher model and the student model at layer \( l_i \), respectively.

    \vspace{-0.2cm}

 \paragraph{Sequence Prediction Loss}
 Finally, except of the distillation losses mentioned above, we also calculate the cross-entropy loss between the output sequence prediction of the student model and the ground truth. This loss is used to guide the student model to learn the correct sequence prediction, which is crucial for the model to perform well on downstream tasks. The loss function is defined as:
 \begin{align}
      \label{eq:ce loss}
      \mathcal{L}_{\text{ce}} = -\sum_{i} y_i \log(\hat{y_{s_i}}),
 \end{align}
 where \( y \) is the ground truth sequence, and \( \hat{y_s} \) is the predicted sequence from the student model.
 
    \vspace{-0.2cm}

 \paragraph{Single Stage Distillation Training}
 To fully harness the complementary strengths of the proposed distillation methods, we integrate the probability distribution loss, layer-wise distillation loss, and the sequence prediction loss into a unified framework for the single stage distillation training. During training, we set their respective weights as follows:
 \begin{align}
    \label{eq:final loss}
    \mathcal{L} = \alpha \cdot \mathcal{L}_{\text{layer}} + \beta \cdot \mathcal{L}_{\text{prob}} + \gamma \cdot \mathcal{L}_{\text{ce}},
 \end{align}
 where \( \alpha, \beta, \gamma \) are hyperparameters that control the relative importance of each loss component.

We will conduct a series of experiments to thoroughly investigate the individual contributions and interactions of these three loss functions. By analyzing their effects in isolation and in combination, we aim to gain deeper insights into how each loss function influences the student model's learning process, the quality of intermediate representations, and the accuracy of final predictions. This will help us understand the specific role of each loss in enhancing the overall performance of the model and ensure that the chosen loss functions are effectively contributing to the optimization process.

This single-stage framework efficiently combines two distillation objectives and one prediction task, allowing gradients from different loss components to flow seamlessly through the student model. The unified loss function not only accelerates convergence during training but also ensures that the student model benefits from both hierarchical feature distillation and global prediction alignment. Furthermore, the proposed method is flexible and can be easily adapted to various neural architectures and tasks by adjusting the weights of the loss components based on task-specific requirements. By integrating these two distillation objectives and the prediction task into a single framework, the student model, \name~, achieves significant performance improvements while maintaining computational efficiency, making this approach highly applicable in real-world scenarios.
    \section{Experiments}
    
    \subsection{Implementation Details}

    We select the TinyLLaVA-Phi-2-SigLIP-3.1B~\cite{zhou2024tinyllava} as the teacher vision-language model model for our experiments. The model's vision encoder is the SigLIP model~\cite{zhai2023siglip}, pre-trained on the WebLi dataset~\cite{chen2022pali} at a resolution of $384\times384$, comprising 400 million parameters. The language model component is Phi-2~\cite{javaheripi2023phi}, featuring 2.7 billion parameters. In the \name~we replace 12.5\%, 25\% and 50\% of the transformer decoder layers in the teacher model with Mamba-2 layers, ensuring an even distribution. As shown in Fig.~\ref{fig:model}, the trainable parameters of the student model are only the Mamba-2 and connector.

    During training, the loss function hyperparameters are set to \( \alpha = \beta = 1.0 \) and \( \gamma = 0 \), indicating that the probability distribution loss and layer-wise distillation loss are assigned equal weights, while the sequence prediction loss is omitted.
    We use a batch size of 64 and optimize the model using the AdamW optimizer with a weight decay of \(0.01\) and momentum parameters \(\beta_1 = 0.9\) and \(\beta_2 = 0.95\). The learning rate is set to \(2 \times 10^{-4}\) and follows a warm-up stable decay schedule, with both the warm-up and decay phases spanning 10\% of the total training steps. Following the TinyLLaVA configuration, we adopt the ShareGPT4V~\cite{chen2024sharegpt4v} SFT dataset for training. This dataset replaces 23K image-text pairs related to the image captioning task in the LLaVA-Mix-665K~\cite{liu2024improvedllava} dataset with an equivalent set of high-quality image-text pairs generated by GPT-4V~\cite{achiam2023gpt4}, ensuring enhanced data quality.
    
    \subsection{Main Results}
    
    \paragraph{Performance Comparison}
    As shown in Tab.~\ref{tab:mllm_comparison}, we show the performance comparison of VLMs across multiple benchmarks, including MME~\cite{fu2023mme}, MMBench~\cite{zhang2023mmbench}, TextVQA~\cite{singh2019textvqa}, GQA~\cite{hudson2019gqa}, MM-Vet~\cite{yu2023mmvet}, ScienceQA~\cite{lu2022sqa}, POPE~\cite{li2023pope}, MMMU~\cite{yue2023mmmu} and VQAv2~\cite{goyal2017vqav2}.
    We present representative VLMs with diverse architectures~\cite{li2023blip,zhu2023minigpt,dai2023instructblip,chen2023shikra,li2023otter,ye2023mplug,laurenccon2024obelics,bai2023qwenvl,liu2024improvedllava} in the table. For comparison, we specifically highlight models have the similar parameter scales~\cite{xu2024llavadi,chu2023mobilevlm,lin2024moe,zhu2024llavaphi,zhou2024tinyllava} with the \name~and those incorporating Mamba~\cite{zhao2024cobra,qiao2024vlmamba,huang2024mlmamba}.
    Firstly, compared to the teacher model TinyLLaVA~\cite{zhou2024tinyllava}, the \name~achieves a 17.6-point improvement on MME. Across all benchmarks, the performance drop remains within 2.6 points, except for MMBench and ScienceQA, which remain within 4.9 points, demonstrating the model's competitive performance. 
    Built upon transformer-based large language models (LLMs)~\cite{chiang2023vicuna,touvron2023llama,bai2023qwen}, the \name~performs comparable on most of the benchmarks, while achieving gains on POPE and MMMU, with performance improvements of 0.2 and 1.6 points,respectively.
    Furthermore, compared to VLMs with similar-scale parameters, our \name~outperforms them across nearly all benchmarks, with notable improvements of 87.7 points on MME and 7.0 points on TextVQA.
    Finally, compared to VLMs incorporating Mamba, our \name~achieves the best performance on most benchmarks, ranking second on TextVQA with only a 0.2-point difference and trailing Cobra~\cite{zhao2024cobra} and ML-Mamba~\cite{huang2024mlmamba} on POPE.
    In summary, these results highlight that our \name~consistently delivers robust and competitive performance across a diverse set of benchmarks, underscoring its effectiveness and strong potential for practical applications.

    \vspace{-0.2cm}

    \paragraph{Inference Speed Comparison}
    We evaluate the inference speed of our \name~against the teacher model TinyLLaVA on the NVIDIA GeForce RTX 3090. As shown in Fig.~\ref{fig:teaser} (b), under the same generated token length setting, the \name~achieves up to $3.6\times$ faster inference compared to the TinyLLaVA with the FlashAttention2~\cite{dao2022flashattention,dao2023flashattention}. In other words, as the generated token length increases, the inference time gap between our \name~and the TinyLLaVA continues to expand. Moreover, a higher hybrid ratio of Mamba-2 layers leads to further improvements in the inference speed.
    This demonstrates the superior efficiency of our \name~during the inference process, making it more suitable for real-world applications.
\begin{table*}[htbp]
    \centering
    \small
    \setlength{\tabcolsep}{1.5pt}
    \renewcommand{\arraystretch}{1.17}
    \begin{tabular}{l l c c c c c c c c c}
        \toprule
        Method & LLM & MME-P & MMB & VQA$^T$ & GQA & MM-Vet & SQA-I & POPE & MMMU & VQAv2 \\
        \midrule
        \rowcolor{gray!20} BLIP-2~\cite{li2023blip} & Vicuna-13B~\cite{chiang2023vicuna} & 1293.8 & - & 42.5 & 41.0 & 22.4 & 61 & 85.3 & - & 41 \\
        \rowcolor{gray!20} MiniGPT-4~\cite{zhu2023minigpt} & Vicuna-7B~\cite{chiang2023vicuna} & 581.7 & 23.0 & - & 32.2 & - & - & - & - & - \\
        \rowcolor{gray!20} InstructBLIP~\cite{dai2023instructblip} & Vicuna-7B~\cite{chiang2023vicuna} & - & 36.0 & 50.1 & 49.2 & 26.2 & 60.5 & - & - & - \\
        \rowcolor{gray!20} InstructBLIP~\cite{dai2023instructblip} & Vicuna-13B~\cite{chiang2023vicuna} & 1212.8 & - & 50.7 & 49.5 & 25.6 & 63.1 & 78.9 & - & - \\
        \rowcolor{gray!20} Shikra~\cite{chen2023shikra} & Vicuna-13B~\cite{chiang2023vicuna} & - & 58.8 & - & - & - & - & - & - & 77.4 \\
        \rowcolor{gray!20} Otter~\cite{li2023otter} & LLaMA-7B~\cite{touvron2023llama} & 1292.3 & 48.3 & - & - & 24.6 & - & - & 32.2 & - \\
        \rowcolor{gray!20} mPLUG-Owl~\cite{ye2023mplug} & LLaMA-7B~\cite{touvron2023llama} & 967.3 & 49.4 & - & - & - & - & - & - & - \\
        \rowcolor{gray!20} IDEFICS-9B~\cite{laurenccon2024obelics} & LLaMA-7B~\cite{touvron2023llama} & - & 48.2 & 25.9 & 38.4 & - & - & - & - & 50.9 \\
        \rowcolor{gray!20} Qwen-VL~\cite{bai2023qwenvl} & Qwen-7B~\cite{bai2023qwen} & - & 38.2 & 63.8 & 59.3 & - & 67.1 & - & - & 78.8 \\
        \rowcolor{gray!20} Qwen-VL-Chat~\cite{bai2023qwenvl} & Qwen-7B~\cite{bai2023qwen} & 1487.5 & 60.6 & 61.5 & 57.5 & - & 68.2 & - & 35.9 & 78.2 \\
        \rowcolor{gray!20} LLaVA-1.5~\cite{liu2024improvedllava} & Vicuna-7B~\cite{chiang2023vicuna} & 1510.7 & 64.3 & 58.2 & 62.0 & 30.5 & 66.8 & 85.9 & - & 78.5 \\
        \rowcolor{gray!20} LLaVA-1.5~\cite{liu2024improvedllava} & Vicuna-13B~\cite{chiang2023vicuna} & 1531.3 & 67.7 & 61.3 & 63.3 & 35.4 & 71.6 & 85.9 & 36.4 & 80 \\
        \rowcolor{blue!10} Teacher-TinyLLaVA~\cite{zhou2024tinyllava} & Phi2-2.7B~\cite{javaheripi2023phi} & 1466.4 & 66.1 & 60.3 & 62.1 & 37.5 & 73.0 & 87.2 & 38.4 & 80.1 \\
        \midrule
\multicolumn{2}{l}{\textit{VLMs with similar scale and Mamba-based ones}} \\
\midrule

        % \midrule
        LLaVA-Phi~\cite{zhu2024llavaphi} & Phi2-2.7B~\cite{javaheripi2023phi} & 1335.1 & 59.8 & 48.6 & - & 28.9 & \underline{68.4} & 85.0 & - & 71.4 \\
        MoE-LLaVA-2.7Bx4~\cite{lin2024moe} & Phi2-2.7B~\cite{javaheripi2023phi} & 1396.4 & \textbf{65.5} & 50.2 & 61.1 & 31.1 & \textbf{68.7} & 85.0 & - & 77.1 \\
        MobileVLM 3B~\cite{chu2023mobilevlm} & MobileLLaMA 2.7B~\cite{chu2023mobilevlm} & 1288.9 & 59.6 & 47.5 & 59.0 & - & 61.0 & 84.9 & - & - \\
        LLaVADI~\cite{xu2024llavadi} & MobileLLaMA 2.7B~\cite{chu2023mobilevlm} & 1376.1 & \underline{62.5} & 50.7 & \underline{61.4} & - & 64.1 & 86.7 & - & - \\
        % \midrule
        % \textit{VLMs with Mamba} \\
        % \midrule
        Cobra~\cite{zhao2024cobra} & Mamba 2.8B~\cite{gu2023mamba} & - & - & \textbf{57.9} & - & - & - & \underline{88.2} & - & 76.9 \\
        VL-Mamba~\cite{qiao2024vlmamba} & Mamba 2.8B~\cite{gu2023mamba} & 1369.6 & 57.0 & 48.9 & 56.2 & 32.6 & 65.4 & 84.4 & - & 76.6 \\
        ML-Mamba~\cite{huang2024mlmamba} & Mamba-2 2.7B~\cite{dao2024mamba2} & - & - & 52.2 & 60.7 & - & - & \textbf{88.3} & - & 75.3 \\
        \textbf{\name~Hybrid-Mamba-12.5\%} & Phi2-2.7B Hybrid Mamba-2 & \underline{1464.9} & 58.7 & 57.5 & 61.2 & \textbf{35.4} & 67.2 & 85.8 & \textbf{38.0} & \underline{78.9} \\
        \textbf{\name~Hybrid-Mamba-25\%} & Phi2-2.7B Hybrid Mamba-2 & \textbf{1484.1} & 61.2 & \underline{57.7} & \textbf{61.5} & \textbf{35.4} & 68.0 & 86.0 & \underline{37.3} & \textbf{79.0} \\
        \bottomrule
    \end{tabular}
    \vspace{-0.2cm}
    \caption{\textbf{Performance comparison of VLMs with our \name~across multiple benchmarks.} We use a \smash{\colorbox{gray!20}{gray}} background to indicate larger LLM-based VLMs and a \smash{\colorbox{blue!10}{blue}} background to highlight the baseline model, TinyLLaVA~\cite{zhou2024tinyllava}. Additionally, we emphasize models with similar parameter scales to our \name, as well as VLMs incorporating Mamba. The best performance for each benchmark column is marked in \textbf{bold}, while the second-best is \underline{underlined}.}
    \vspace{-0.1cm}
    \label{tab:mllm_comparison}
\end{table*}
\begin{table*}[htbp]
    \centering
    \small
    \setlength{\tabcolsep}{5.6pt}
    \renewcommand{\arraystretch}{1.1}
    \begin{tabular}{l|cccccccccc}
        \toprule
        Method & MME-P & MMB & VQA$^T$ & GQA & MM-Vet & SQA-I & POPE & MMMU & VQAv2 & AVG \\
        \midrule
        \name~Hybrid-Mamba-12.5\% & \underline{1464.9} & 58.7 & \underline{57.5} & \underline{61.2} & \textbf{35.4} & \underline{67.2} & \underline{85.8} & \textbf{38.0} & \underline{78.9} & \underline{61.8} \\
        \name~Hybrid-Mamba-25\% & \textbf{1484.1} & \textbf{61.2} & \textbf{57.7} & \textbf{61.5} & \textbf{35.4} & \textbf{68.0} & \textbf{86.0} & \underline{37.3} & \textbf{79.0} & \textbf{62.3} \\
        \name~Hybrid-Mamba-50\% & 1427.8 & \underline{59.4} & 54.3 & 60.8 & 30.5 & 64.9 & \underline{85.8} & 34.8 & 78.6 & 60.1 \\
        \bottomrule
    \end{tabular}
    \vspace{-0.2cm}
    \caption{Performance comparison for different \name~hybrid the Mamba-2 configurations. The best performance for each benchmark column is marked in \textbf{bold}, while the second-best is \underline{underlined}.}
    \label{tab:hybrid_ratio}
    % \vspace{-0.2cm}
\end{table*}
    
    \vspace{-0.2cm}
    \paragraph{Memory Usage Comparison}
    We further compare the GPU memory usage of our \name~with the TinyLLaVA on the NVIDIA GeForce RTX 3090. As illustrated in Fig.~\ref{fig:teaser} (c), our \name~demonstrates a substantially lower memory footprint compared to the TinyLLaVA, achieving a peak reduction of 27.5\% at a token length of 16,384. Notably, the TinyLLaVA encounters an out-of-memory error at a token length of 32,768, whereas our \name~continues to run without issue.
    This reduction in memory usage is attributed to the \name's optimized architecture, which effectively balances computational efficiency and performance, making it more suitable for deployment on resource-constrained devices.
    
    \subsection{Ablation Study}

    \begin{table*}[htbp]
        \centering
        \small
        \setlength{\tabcolsep}{6.9pt}
        \renewcommand{\arraystretch}{1.1}
        \begin{tabular}{l|cccccccccc}
            \toprule
            Mamba-2 Layer Position & MME-P & MMB & VQA$^T$ & GQA & MM-Vet & SQA-I & POPE & MMMU & VQAv2 & AVG \\
            \midrule
            Evenly Distributed & \textbf{1484.1} & \textbf{61.2} & \textbf{57.7} & \textbf{61.5} & \textbf{35.4} & \textbf{68.0} & \underline{86.0} & \textbf{37.3} & \textbf{79.0} & \textbf{62.3} \\
            All at the Beginning & \underline{1406.5} & \underline{60.8} & \underline{56.1} & \underline{61.2} & \underline{32.9} & \underline{64.3} & \textbf{86.1} & \underline{33.0} & \underline{78.8} & \underline{60.4} \\
            All in the Middle & 1392.9 & 54.0 & 53.0 & 60.0 & 32.4 & 63.3 & 85.0 & 34.9 & 77.9 & 58.9 \\
            \bottomrule
        \end{tabular}
    \vspace{-0.1cm}
        \caption{Effect of different Mamba-2 layer positions. The best performance for each benchmark column is marked in \textbf{bold}, while the second-best is \underline{underlined}.}
    \vspace{-0.1cm}
        \label{tab:hybrid_position}
    \end{table*}
    \begin{table*}[htbp]
        \centering
        \setlength{\tabcolsep}{7.3pt}
        \small
        \renewcommand{\arraystretch}{1.2}
        \begin{tabular}{l|cccccccccc}
            \toprule
            Distillation Loss & MME-P & MMB & VQA$^T$ & GQA & MM-Vet & SQA-I & POPE & MMMU & VQAv2 & AVG \\
            \midrule
            $L_{ce}$ & 1284.0 & 56.4 & 45.3 & 57.5 & 29.0 & 59.6 & \textbf{86.0} & 28.0 & 74.8 & 55.6 \\
            $L_{layer}$ & \underline{1430.7} & \textbf{61.9}& 55.7 & 60.7 & 33.2 & 66.9 & 84.7 & 33.0 & 78.8 & 60.7 \\
            $L_{prob}$ & 1413.4 & \underline{61.5} & \underline{56.8} & \underline{61.0} & \underline{33.3} & \underline{67.0} & 85.7 & \textbf{37.7} & 78.8 & \underline{61.4} \\
            $L_{prob} + L_{layer}$ & \textbf{1484.1} & 61.2 & \textbf{57.7} & \textbf{61.5} & \textbf{35.4} & \textbf{68.0} & \textbf{86.0} & \underline{37.3} &  \underline{79.0} & \textbf{62.3} \\
            $L_{prob} + L_{layer} + L_{ce}$ & 1449.8 & 60.9 & 57.0 & 61.0 & 32.6 & 67.6 & 85.8 & 35.4 & \textbf{79.1} & 61.3 \\
            \bottomrule
        \end{tabular}
    \vspace{-0.1cm}
        \caption{Comparison of different distillation loss functions. The best performance for each benchmark column is marked in \textbf{bold}, while the second-best is \underline{underlined}.}
        \label{tab:distillation_loss}
    \end{table*}
    
    \paragraph{Mamba-2 Hybridization Ratio} We examine the impact of the Mamba-2 hybridization ratio by varying the proportion of Mamba-2 layers (12.5\%, 25\%, and 50\%) and evaluating performance across eight multi-modal benchmarks. As shown in Tab.~\ref{tab:hybrid_ratio}, the 25\% ratio achieves the highest average score, surpassing the 50\% ratio by 2.2 points, indicating that excessive Mamba-2 layers may weaken global dependency modeling. The 12.5\% ratio achieves the best performance on MM-Vet and MMMU but falls slightly behind overall, scoring 0.5 points lower than the 25\% ratio. This finding highlights the importance of balancing Mamba-2 and transformer layers to optimize performance across diverse tasks. While a higher Mamba-2 ratio (50\%) may hinder the model's ability to capture long-range dependencies, a lower ratio (12.5\%) retains transformer advantages but may not fully leverage Mamba-2’s benefits. 
    
    \vspace{-0.2cm}

    \paragraph{Mamba-2 Hybridization Layer Position} We further investigate the impact of Mamba-2 layer placement on the performance. Specifically, we replace the transformer decoder layers with Mamba-2 layers in four configurations: all at the beginning, all in the middle, all at the end, and evenly distributed. Notably, the all-at-the-end configuration fails to enable effective distillation, resulting in incoherent responses. As shown in Tab.~\ref{tab:hybrid_position}, the evenly distributed configuration achieves the highest performance across all benchmarks, surpassing the all-at-the-beginning and all-in-the-middle configurations by 1.9 and 3.4 points on the average score, respectively. These results highlight the importance of evenly integrating Mamba-2 layers throughout the model to optimize performance.

    \vspace{-0.2cm}

    \paragraph{Distillation Loss} As mentioned in Sec~\ref{sec:distillation}, we employ three distillation losses: probability distribution loss $L_{prob}$, layer-wise distillation loss $L_{layer}$, and sequence prediction loss $L_{ce}$. To investigate the impact of each loss on the performance of our \name, we conduct an ablation study, as shown in Tab.~\ref{tab:distillation_loss}. Initially, we use the three losses individually. The results show that $L_{prob}$ significantly improves performance over $L_{ce}$, with a 5.8-point increase in average score. Adding $L_{layer}$ further boosts performance, yielding the highest average score when combining $L_{prob}$ and $L_{layer}$, which is 0.9 points higher than using $L_{prob}$ alone. This indicates that both probability distribution alignment and layer-wise feature matching contribute positively to knowledge transfer. However, when $L_{ce}$ is reintroduced alongside $L_{prob}$ and $L_{layer}$, we observe a slight 1.0-point drop, suggesting that direct supervision from $L_{ce}$ may interfere with the distillation process.

\section{Limitations}
Despite its advantages, \name~has several limitations. While initializing Mamba-2 with pre-trained attention weights aids convergence, it may not fully leverage its implicit state representations. This could be improved through tailored initialization strategies, such as gradient matching or an additional pretraining phase. Additionally, due to the limited GPU resources available in this study, we have not explored the performance of our model at larger scales. With access to more computational resources, future work can systematically investigate optimal Mamba-2 integration ratios and conduct hybrid experiments on larger VLMs to evaluate scalability and performance. Addressing these challenges will further improve the efficiency and applicability of hybrid architectures in large-scale VLMs.

\section{Conclusion}
We propose a hybrid model, \name, that enhances a pre-trained vision-language model (VLM) by replacing a portion of its transformer decoder layers with Mamba-2 layers. This design leverages the efficiency of RNN-inspired architectures while preserving the expressiveness of transformers. By initializing Mamba-2 layers with attention weights and employing a single-stage distillation process, we improve both convergence speed and overall performance. Notably, our model is trained using only four NVIDIA GeForce RTX 3090 GPUs, demonstrating its computational efficiency.
Extensive evaluations show that our approach not only achieves competitive accuracy but also significantly enhances inference speed and reduces GPU memory consumption. These results highlight the potential of hybrid architectures to strike a balance between efficiency and expressiveness. Furthermore, the efficiency of our model in both fine-tuning and inference makes it a cost-effective and scalable solution for deploying large-scale VLMs. This approach enables more practical, resource-efficient deployment of VLMs in real-world applications, reducing computational costs while maintaining high performance.
{
    \small

% \clearpage % 确保所有浮动体在继续正文前被放置
    \bibliographystyle{ieeenat_fullname}
    \bibliography{main}
}

\end{document}